# Parsimonious neural networks learn interpretable physical laws

*Saaketh Desai and Alejandro Strachan*
*School of Materials Engineering and Birck Nanotechnology Center*
*Purdue University, West Lafayette, Indiana, 47907 USA*

Machine learning is playing an increasing role in the physical sciences and significant progress has been made towards embedding domain knowledge into models. Less explored is its use to discover interpretable physical laws from data. We propose parsimonious neural networks (PNNs) that combine neural networks with evolutionary optimization to find models that balance accuracy with parsimony. The power and versatility of the approach is demonstrated by developing models for classical mechanics and to predict the melting temperature of materials from fundamental properties. In the first example, the resulting PNNs are easily interpretable as Newton's second law, expressed as a non-trivial time integrator that exhibits time-reversibility and conserves energy, where the parsimony is critical to extract underlying symmetries from the data. In the second case, the PNNs not only find the celebrated Lindemann melting law, but also new relationships that outperform it in the pareto sense of parsimony vs. accuracy.

Machine learning (ML) can provide predictive models in applications where data is plentiful and the underlying governing laws are unknown [1–3]. These approaches are also playing an increasing role in the physical sciences where data is generally limited but underlying laws (sometimes approximate) exist [4–9]. For example, ML-based constitutive models are being used in electronic structure calculations [10] and molecular dynamics (MD) simulations [11–13]. One of the major drawbacks of the use of ML in the physical sciences is that models often do not *learn* the underlying physics of the system at hand, such as constraints or symmetries, limiting their ability to generalize. In addition, most ML models lack interpretability. That is, ML approaches generally neither learn physics nor can they explain their predictions. In many fields, these limitations are compensated by copious amounts of data, but this is often not possible in areas such as materials science where acquiring data is expensive and time consuming. To tackle this challenge, progress has been made towards using knowledge (even partial) of underlying physics to improve the accuracy of models and/or reduce the amount of data required during training [14–16]. Less explored is the use of ML for scientific discovery, i.e. extracting physical laws from observational data, see Refs. [17–19] for recent progress. In this letter, we combine neural networks (NNs) with stochastic optimization to find models that balance accuracy and parsimony and apply them to learn, solely from observational data, the dynamics of a particle under a highly non-linear potential, and expressions to predict the melting temperature of materials in terms of fundamental properties. Our hypothesis is that the requirement of parsimony will result in the discovery of the physical laws underlying the problem and result in interpretability and improved generalizability. We find that the resulting descriptions are indeed interpretable and provide insight into the system of interest. In the case of particle dynamics, the learned models satisfy non-trivial underlying symmetries embedded in the data which increases the applicability the parsimonious neural networks (PNNs) over generic NN models. Stochastic optimization has been previously used in conjunction with backpropagation to improve robustness or minimize overfitting in models [20–26], this work extends these ideas to finding parsimonious models from data to learn physics.

The power of physics-based ML is well documented and remains an active area of research. Neural networks have been used to both parametrize and solve differential equations such as Navier Stokes [14,15] and Hamilton's equations of motion [27]. Recurrent architectures have also shown promise in predicting the time evolution of systems [28,29]. These examples focus on using prior knowledge of the underlying physics to guide the model, often as numerical constraints, or by using the underlying physics to numerically solve equations with variables predicted by the ML algorithms. In contrast, we are interested in learning physics, including the associated numerical solutions, directly from data, without prior knowledge. Pioneering work along these lines used symbolic regression methods, enhanced by matching partial derivatives to identify invariants [17], or using dimensionality reduction and other symmetry-identifying methods to aid equation discovery [30]. These approaches also consider the tradeoff between parsimony and accuracy to develop simple models that describe the data well. On the other hand, neural networks such as time-lagged autoencoders have also proven useful at extracting laws that govern the time evolution of systems from data [31], where the encoder networks attempt to learn features relevant to the problem. Advances here have considered networks with custom activation functions whose weights decide the functional form of the equation [32,33]. Lastly, other approaches to learning physics from data have focused on discovering partial differential equations directly from data, either using a library of candidate derivatives coupled with linear regression [18,19], or using neural networks coupled with genetic algorithms to identify differential equations from an incomplete library [34]. We build on and extend these ideas to propose PNNs, models designed to balance parsimony with accuracy in describing the training data. The PNN approach allows complex compositions of functions via the use of neural networks, while balancing for parsimony using genetic algorithms. As will be shown with two examples, our approach is quite versatile and applicable to situations where an underlying differential equation may not exist. We first apply PNNs to learn the equations of motion that govern the Hamiltonian dynamics of a particle under a highly non-linear external potential with and without friction. Our hypothesis is that by requiring parsimony (e.g. minimizing adjustable parameters and favoring linear relationships between variables) the resulting model will not only be easily interpretable but also will be forced to tease out the symmetries of the problem. We find that the resulting PNN not only lends itself to interpretation (as Newton's laws) but also provides a significantly more accurate description of the dynamics of the particle when applied iteratively as compared to a flexible feed forward neural network. The resulting PNNs conserve energy and are time reversible, i.e. they learn non-trivial symmetries embedded in the data but not explicitly provided. This versatility and the generalizability of PNNs is demonstrated with a second, radically different, example: discovering models to predict the melting temperature of materials from atomic and crystal properties. By varying the relative importance of parsimony and accuracy in the genetic optimization, we discover a family of melting laws that include the celebrated Lindemann law [35]. Quite remarkably the Lindemann law, proposed in 1910, is near (but not on) the resulting pareto front.

# Results

**Discovering integration schemes from data.** As a first example, we consider the dynamics of a particle under an external Lennard-Jones (LJ) potential with and without friction. In both cases the training data is obtained from accurate numerical trajectories with various totals energies. The input and output data are positions and velocities at a given time and one timestep ahead, respectively (this timestep is ten times what was used to generate the underlying trajectories). The numerical data was divided into training and validation sets and an independent testing set was generated at a different energy, see Methods and section S1 of the Supplementary Material (SM). The input data in this example has been generated numerically for convenience but could have been obtained experimentally, as will be shown in the second example. Before describing the PNN model, we establish a baseline by training a standard feed forward neural network (FFNN) on our data for the case without friction. The details of this architecture can be found in section S2 of the SM and can be accessed for online interactive computing on nanoHUB [36]. We find the FFNN to be capable of matching the training/validation/test data well, with root mean squared errors (RMSEs) across all sets on the order of $10^{-5}$ in LJ units for both positions and velocities (see Figure S1 in SM). However, the network has poor predictive power. Using it iteratively to find the temporal evolution of the particle results in significant drifts in total energy over time, and a lack of time reversibility. Reversibility is judged by applying the network sequentially 1,000 times, followed by time reversal (changing the sign of the particle's velocity) and applying the NN for a second set of 1,000 steps. We find that deeper architectures do not improve the RMSE, reversibility or energy conservation. Needless to say, these FFNNs are not interpretable. These results highlight the difficulty of the problem at hand. Hamilton's equations for classical mechanics represent a stiff set of differential equations and small errors in each timestep accumulate rapidly resulting in diverging trajectories. Prior attempts to address this challenge in the context of discovering differential equations explicitly trained models for multiple steps using a recurrent architecture [37]. The resulting models are interpretable and improve accuracy over the number of steps used in training the recurrent network but accumulate relatively high errors over multiple steps. In contrast, we are interested in solutions stable over timescales far greater than those typically accessed by current recurrent architectures, while also favoring the discovery of constants relevant to the physical problem. Finding such models is non-trivial and the development of algorithms to integrate equations of motion with good energy conservation and time reversibility has a rich history [38–42]. An example of such algorithms is the popular Verlet family of integrators [38,39] that are both reversible and symplectic [43]; their theoretical justification lies in Trotter's theorem [44].

**Parsimonious neural networks**. Having established the shortcomings of the state-of-the-art neural networks, we introduce parsimonious neural networks (PNN). We begin with a generic neural network shown in Figure 1 and use genetic algorithms to find models with controllable parsimony. In this first example, the neural network consists of three hidden layers and an output layer with two values, the position and velocity of the particle one timestep ahead of the inputs. Each hidden layer has two neurons, and the central layer includes an additional force sub-net, a network pre-trained to predict the force on the atom given its position. The use of a pre-trained force sub-net is motivated by the prior success of neural networks in predicting interatomic forces in a wide variety of materials significantly more complex than our example [45–47]. In addition, our focus is on learning classical dynamics and the use of a force sub-net only incorporates the physical insight that the force is an important quantity. As a second baseline, we trained a feed forward NN including a pre-trained force sub-net. This second network's performance is as poor as the previous baseline feed-forward network, see section S3 in the SM for details. This shows that adding the information about the force is not the key to the development of accurate models for classical mechanics, parsimony is.

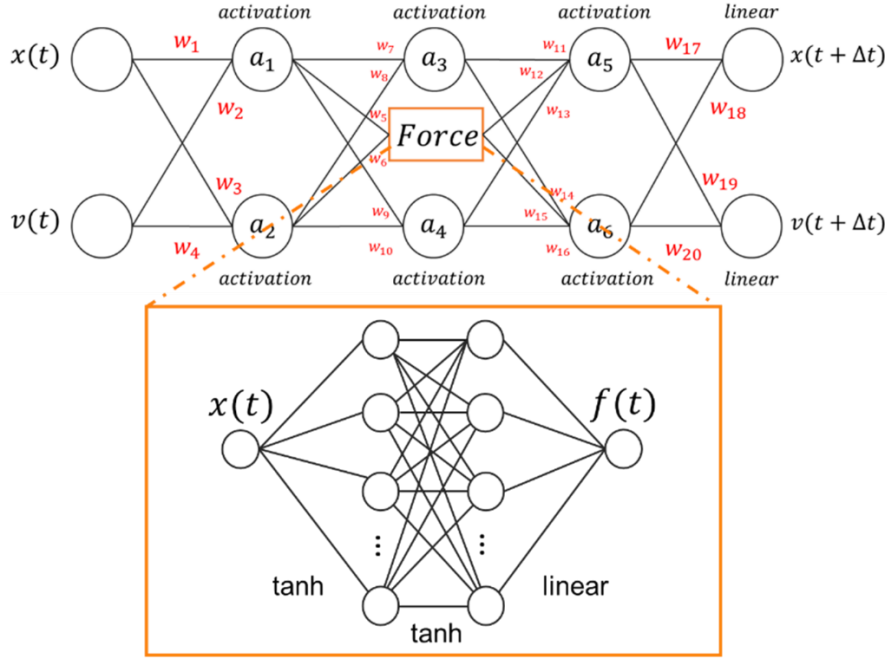

*Figure 1: Neural network used as the starting point to find the parsimonious neural network as the network that explains the data in the simplest manner possible. The force sub-network is highlighted in orange and is fed into the neural network as a pre-trained model, whose weights are subsequently kept fixed throughout*

The starting neural network provides a highly flexible mapping from input positions and velocities to output positions and velocities and PNNs seeks to balance simplicity and accuracy in reproducing the training data. This is an optimization problem in the space of functions spanned by the possible activations and weights of the network. We consider four possible activation functions in this example: linear, rectified linear unit (relu), hyperbolic tangent (tanh), and exponential linear unit (elu). The weights connecting the artificial neurons can be either *fixed* or *trainable*, with the fixed category allowing the following values: 0, ½, 1, 2, $\frac{\Delta t}{2}$, $\Delta t$, and $2\Delta t$, with $\Delta t$ the timestep separating the inputs and outputs. This is motivated by the fact that physical laws often involve integer or simple fractional coefficients and that the timestep represents important information. Our network has twenty weights (each with eight possible settings) and six activation functions to optimize, see Figure 1 (top panel). A brute force approach to finding a PNN model would require training ~$10^{21}$ neural networks, an impossible computational task even for the relatively small networks here. We thus resort to evolutionary optimization, using a genetic algorithm to discover models that balance accuracy and parsimony.

PNNs use an objective function that includes measures of the accuracy in the test set and parsimony. The latter term favors: i) linear activation functions over non-linear ones, and ii) non-trainable weights with simple values over optimizable weights. The objective function for the genetic optimization is defined as:

$$F = f_1(E_{test}) + p\left(\Sigma_{i=1}^{N_N} w_i^2 + \Sigma_{j=1}^{N_w} f_2(w_j)\right) \tag{1}$$

where $E_{test}$ represents the mean squared error of the trained PNN on the testing set and $f_1$ is a logarithmic function that converts the wide range errors into a scale comparable to the parsimony terms, see section S4 of the SM. The second term runs over the $N_N$ neurons in the network and is designed to favor simple activation functions. The *linear*, *relu*, *tanh* and *elu* activation functions are assigned scores of $w_i = $ 0, 1, 2 and 3, respectively. The third term runs over the network weights and biases ($N_w$) and favors fixed, simple weights over trainable ones. A fixed weight value of 0 is assigned a score of 0, while other fixed weights are assigned the score 1, and a trainable weight is assigned a score of 2. $p$ is a parsimony coefficient that determines the relative importance of parsimony and accuracy. We use the DEAP package for the evolutionary optimization [48] and Keras [49] to train individual networks, see Methods. We note that our approach is similar in spirit to recent work combining genetic algorithms with neural networks to discover partial differential equations [34], but PNNs are more versatile in terms of parsimony, composition of functions, and are applicable to situations where an underlying differential equation may not exist, as we will see in the second example discussed in this paper. We also note that evolutionary optimization is not the only way to achieve parsimony. For example, one could include hidden layers containing a library of possible activation functions and use sparsity to prune unnecessary activations. This has recently been used to discover simple kinematics equations [33]. An advantage of this approach over ours is simplicity and computational expedience since such networks can be trained using backpropagation alone. However, the evolutionary approach used in PNNs offers significant advantages including a more efficient exploration of function space and avoiding local minima, flexibility in the definition of parsimony, and composition of functions via the neural network.

The PNNs resulting from a genetic optimization with $p = 1$ reproduce the training, validation and testing data more accurately than the architecturally complex FFNNs. Figure 2(a) compares the RMSE for positions and velocities from the optimal PNN (denoted PNN1) to the FFNN. Remarkably, PNN1 also results in excellent long-term energy conservation and time reversibility, evaluated using the same procedure as before. Figures 2(b) and 2(c) compare the total energy and trajectories generated by PNN1, the FFNN, and the state-of-the-art velocity Verlet integrator. We see that PNN1 *learns* both time-reversibility and that total energy is a constant of motion. This is in stark contrast to the physics-agnostic FFNN and even naïve physics-based models like a first order Euler integration. A few of the top ranked PNNs perform similarly to PNN1 and they will be discussed in Section S7 of the SM.

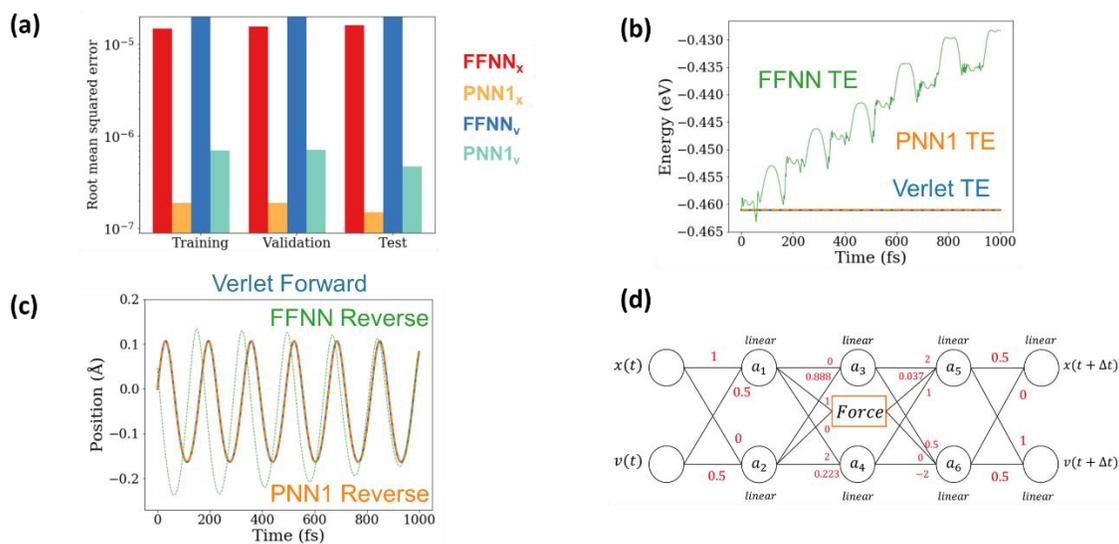

*Figure 2: (a) PNN model 1 RMSEs on the training/validation/test sets compared to the feed forward network (b) We see that energy conservation between PNN1 and the verlet integrator is comparable (TE: total energy) (c) Forward and reverse trajectories generated by PNN1 show good reversibility (d) A visualization of PNN model 1 found by the genetic algorithm, attempting to predict positions and velocities one step ahead*

Having established that the PNNs *learn* the physics of the system and result in stable and accurate integrators, we now explore their interpretability in the hope of finding out how time-reversibility and energy conservation are achieved. In short: can the PNNs *teach* us what they learned? We find that the PNNs discover simple models, with many weights taking fixed values (including zero) and all activations functions taking the simplest possible alternative (linear functions). As an example, the parameters corresponding to PNN1 are shown in Figure 2(d), and other PNNs with comparable (but higher) objective functions are shown in Section S6 of the SM. This simplicity allows us to trivially obtain position and velocity update equations. The equations of motion learned by PNN1, rewritten in terms of relevant quantities such as timestep and mass, are:

$$x(t + \Delta t) = x(t) + 1.0001 \, v(t)\Delta t + 0.9997 \, \frac{1}{2} F\left(x(t) + v(t)\frac{\Delta t}{2}\right)\frac{\Delta t^2}{m} \tag{2}$$

$$v(t + \Delta t) = v(t) + 0.9997 F\left(x(t) + v(t)\frac{\Delta t}{2}\right)\frac{\Delta t}{m} \tag{3}$$

Inspecting Figure 2(d) and Eqs. (2,3) we find that PNN1 achieves time-reversibility by evaluating the force at the midpoint between inputs and outputs, this central force evaluation is key to many advanced numerical methods. In fact, PNN1 represents the position Verlet algorithm [39] except that the NN training makes an error in the mass of approximately 3 in 10,000. This algorithm is both reversible and symplectic, i.e. it conserves volume in phase space. The small error in mass seems to originate from the small inaccuracies of the force sub-net to describe the Lennard -Jones potential, see Section S6 of the SM.

The genetic optimization provides an ensemble of models and inspecting slightly sub-optimal ones provides interesting insights. The SM provides the equations of motion predicted by PNN2 and 3. These are similarly interpretable and, quite remarkably, they also learn to evaluate the force at the half-step. They represent a slightly inaccurate version of the position Verlet algorithm with minor energy drifts due to a slight asymmetry in effective mass in the position and velocity update equations. Finally, changing the parsimony parameter $p$ in the objective function, allows us to generate a family of models with different tradeoffs between accuracy and parsimony; see Figure

S8 in the SM. Interestingly, we find models that reproduce the training and testing data more accurately than PNN 1 and Verlet. However, these models are not time reversible and their energy conservation is worse than PNN1, see Section S7 in the SM, stressing the importance of parsimony.

Along similar lines, we tested the ability of the PNNs to discover the physics governing a damped dynamical system, see Methods. The equations learned by the top PNN, with $\gamma$ the damping constant, are:

$$x(t + \Delta t) = x(t) + 1.0008\, v(t) * \left(\Delta t - \frac{\gamma \Delta t^2}{2m}\right) + 0.9991 \frac{1}{2} F\left(x(t) + v(t)\frac{\Delta t}{2}\right) \frac{\Delta t^2}{m} \quad (4)$$

$$v(t + \Delta t) = \left(1 - \frac{\gamma \Delta t}{m}\right) v(t) + 1.0002\, F\left(x(t) + v(t)\frac{\Delta t}{2}\right) \frac{\Delta t}{m} \quad (5)$$

In this second example, PNNs learn classical mechanics, the fact that the frictional force is proportional to negative the velocity, and discover the same stable integrators based on the position Verlet method, all from the observational data.

We consider the emergence of Verlet style integrators from data remarkable. This family of integrators is the preferred choice for molecular dynamics simulations due to its stability. Unlike other algorithms such as the Runge-Kutta family or the first order Euler method, Verlet integrators are symplectic and time reversible [50]. This class of integrators has been long known, and proposed independently by several researchers over decades (see Ref. [50] for a review), but a detailed understanding of their properties and their justification from Trotter's theorem are relatively modern [39]. Importantly, we find more complex models that reproduce the data more accurately than PNN1 but do not exhibit time reversibility nor conserve energy. This shows that parsimony is critical to learn models that can provide insight into the physical system at hand and for generalizability. We stress that the equations of motion and an advanced integrator were obtained from observational data of the motion of a particle and the force-displacement relationship alone. We believe that, at the expense of computational cost, the force sub-net could be learned together with the integrators (effectively learning the acceleration) from large-enough dynamical datasets. This assertion is based on the observation that on fixing some of the network parameters that result in a Verlet integrator, the remaining parameters and the force subnet can be learned from the observational data used above, see section S7 of the SM.

**Melting temperature laws.** To demonstrate the versatility and generalizability of PNNs, we now apply them to discover melting laws from experimental data. Our goal is to predict the melting temperature of materials from fundamental atomic and crystal properties. To do this, we collected experimental melting temperatures for 218 materials (including oxides, metals, and other single elements crystals) as well as fundamental physical quantities including: bulk modulus $K$, shear modulus $G$, density $\rho$, a characteristic atomic distance $a$ (the cube root of the volume per atom), and mean atomic mass $m$.

Before feeding this data to PNNs, we perform a standard dimensionality analysis to use dimensionless inputs and output. For convenience we first define an effective sound speed, $v_m$, from density and elasticity moduli, see SM. From these fundamental quantities, we define four independent quantities with the dimensions of temperature:

$$\theta_0 = \frac{\hbar v_m}{k_b a} \tag{6}$$

$$\theta_1 = \frac{\hbar^2}{m a^2 k_b} \tag{7}$$

$$\theta_2 = \frac{a^3 G}{k_b} \tag{8}$$

$$\theta_3 = \frac{a^3 K}{k_b} \tag{9}$$

where is $\hbar$ Planck's constant and $k_b$ is the Boltzmann's constant. All variables have physical meanings, for example, $\theta_0$ is proportional to the Debye temperature. The inputs to the PNNs are the last three quantities normalized by $\theta_0$ and the output melting temperature is also normalized by $\theta_0$. Additional details on the preprocessing steps as well as network architecture, including custom activations, can be found in the Section S8 of the SM.

Armed with dimensionless inputs and outputs, we use PNNs to discover melting laws. Varying the parsimony parameter in the objective function, Eq. 1, results in a family of melting laws. These models are presented in Figure 3 in terms of their accuracy with respect to the testing set and their complexity. The latter is defined as the sum of the second and third terms of the objective function

Eq. 1, i.e., the sum of the activation function and weight terms. PNN models represent various tradeoffs between accuracy and parsimony from which we can define a pareto front of optimal models (see dashed line).

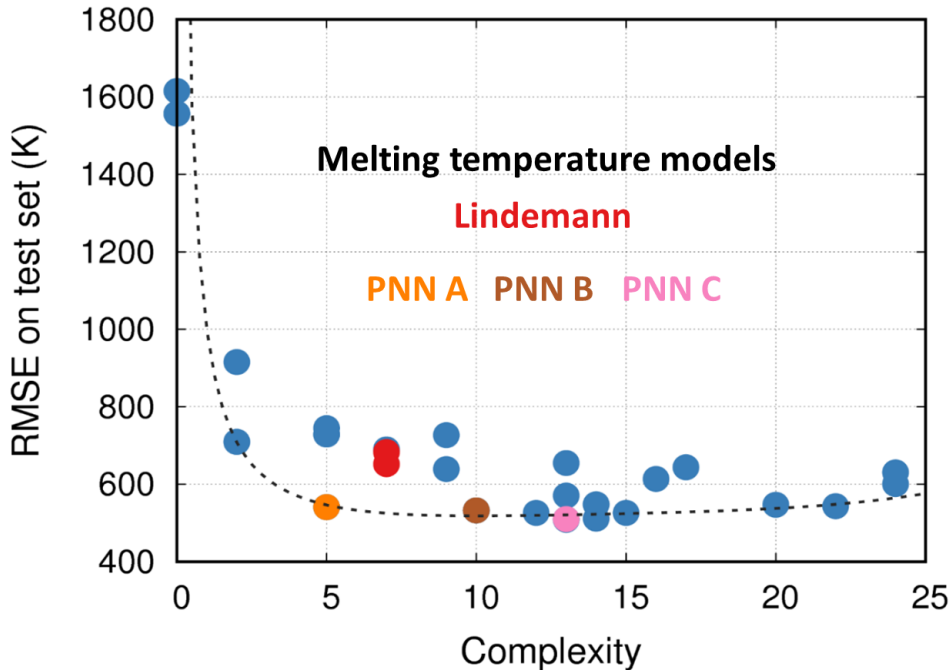

*Figure 3: Melting laws discovered by PNNs. The red points show the celebrated Lindemann law, while the blue points show other models discovered. The black dotted line denotes the pareto front of models, with some of the models performing better than the Lindemann law while also being simpler. Three models are highlighted and labeled.*

The PNN approach finds several simple yet accurate expressions. The simplest non-trivial relationship is given by PNN A, it approximates the melting temperature as proportional to the Debye temperature:

$$T_m^{PNN\ A} = 21.8671\ \theta_0 \tag{10}$$

This makes physical sense as the Debye temperature is related to the characteristic atomic vibrational frequencies and stiffer and stronger bonds tend to lead to higher Debye and melting temperatures. Next in complexity, PNN B adds a correction proportional to the shear modulus:

$$T_m^{PNN\ B} = 17.553\ \theta_0 + 0.001985\ \theta_2 \tag{11}$$

This is also physically sensical as shear stiffness is closely related to melting. This fact is captured by the classic Born instability criteria [51] that associates melting to loss of shear stiffness. Just above PNN B in complexity, PNN finds the celebrated Lindemann melting law [35]:

$$T_m^{lind} = C \frac{\theta_0^2}{\theta_1} = \frac{k_b}{9\hbar^2} f^2 a^2 m T_D^2 \tag{12}$$

Written in its classical form in the third term; here $T_D$ is the debye temperature of the material and $f$ and $C$ are an empirical constant. Remarkably, this law, derived using physical intuition in 1910, is very close to, but not on, the optimal pareto front in accuracy-complexity space. For completeness, we describe the model with the lowest RMS error, PNN C predicts the melting temperature as:

$$T_m^{PNN\ C} = 11.9034\ \theta_0 + 0.000499\ \theta_3 + 0.00796 \frac{\theta_0^2}{\theta_1} \tag{13}$$

Quite interestingly, this model combines the Lindemann expression with Debye temperature and bulk (not shear) modulus. This combination is not surprising given the expressions above, but the selection of bulk over shear modulus is not clear at this point and should be explored further.

In summary, we proposed parsimonious neural networks that are capable of learning interpretable physics models from data; importantly, they can extract underlying symmetries in the problem at hand and provide physical insight. The combination of genetic optimization with neural networks enables PNNs to explore a large function space and obviate the need for estimating numerical derivatives or matching a library of candidate functions, as was done in prior efforts [17–19]. Additionally, PNNs perform complex composition of functions in contrast to sparse regression, which combine functions linearly. As opposed to most efforts attempting to discover differential equations, such as DLGA-PDE, PNN can discover laws in situations where an underlying differential equation may not exist. The state-of-art solutions PNNs provide two quite different problems attest to the power and versality of the approach.

From data describing the classical evolution of a particle in an external potential, the PNN produces integration schemes that are accurate, conserve energy and satisfy time reversibility. Furthermore, they can be easily interpretable as discrete versions of Newton's equations of motion. Quite interestingly, the PNNs learn the non-trivial need to evaluate the force at the half step for time reversibility. The optimization could have learned the first order Runge-Kutta algorithm, which is not reversible, but it favored central-difference based integrators. Furthermore, parsimony favors Verlet-type integrators over more complex expressions that describe the training data more accurately but do not exhibit good stability. We note that other high-order integrators are not compatible with our initial network, but these can easily be incorporated by starting with a more complex network. As discussed above, the resulting algorithms would not come as a surprise to experts in molecular dynamics simulations as this community has developed, over decades, accurate algorithms to integrate Newton's equations of motion. The fact that such knowledge and algorithms can be extracted automatically from observational data has, however, deep implications in other problems and fields. This is confirmed with a second example that shows the ability of PNNs to extract interpretable melting laws from experimental data. We discover a family of expressions with varying tradeoffs between accuracy vs. parsimony and our results show that the widely used Lindeman law, proposed over a century ago, is remarkably close to the pareto front but we find PNNs that outperform it. The PNN models highlight the relationships between melting and a materials Debye's temperature as well as shear moduli, providing insight into the processes that determine melting.

### Acknowledgements

Partial support from the Network for Computational Nanotechnology, Grant EEC-1227110, is acknowledged.

## Methods

**Training data**. To discover integration schemes, training data was generated using molecular dynamics simulations under an NVE ensemble, using the velocity Verlet scheme with a short timestep (about a tenth of what is required for accurate integration), see section S1 of the SM for additional details. Training and validation data are obtained from trajectories with four different total energies, 20% of which is used as a validation set. Our test set is a separate trajectory with a different total energy. For the damped dynamics cases, a frictional force proportional to negative the velocity is added, with frictional coefficient $\gamma = 0.004$ eVps/Å$^2$. To discover novel melting laws, we queried the Pymatgen and Wolfram alpha databases for experimental melting temperatures. We obtained fundamental material properties such as volume and shear modulus by querying the Materials Project. Additional details are provided in section S8 of the SM.

**Evolutionary optimization**. We used populations of 200 and 500 individuals and a two-point crossover scheme and random mutations to evolve the population (weights and activation functions) [52]. For each generation, individual networks in the population are trained using backpropagation using the same protocols as for the feed-forward networks; only adjustable weights are optimized in this operation. The populations were evolved over 50 generations, additional details of the genetic algorithm are included in section S5 of the SM.

References

[1] K. Simonyan and A. Zisserman, *Very Deep Convolutional Networks for Large-Scale Image Recognition*, ArXiv Preprint ArXiv:1409.1556 (2014).
[2] A. Krizhevsky, I. Sutskever, and G. E. Hinton, *ImageNet Classification with Deep Convolutional Neural Networks*, in *Advances in Neural Information Processing Systems 25*, edited by F. Pereira, C. J. C. Burges, L. Bottou, and K. Q. Weinberger (Curran Associates, Inc., 2012), pp. 1097–1105.
[3] Y. Bengio, R. Ducharme, P. Vincent, and C. Jauvin, *A Neural Probabilistic Language Model*, Journal of Machine Learning Research **3**, 1137 (2003).
[4] J. Carrasquilla and R. G. Melko, *Machine Learning Phases of Matter*, Nature Physics **13**, 431 (2017).
[5] A. W. Senior, R. Evans, J. Jumper, J. Kirkpatrick, L. Sifre, T. Green, C. Qin, A. Žídek, A. W. R. Nelson, A. Bridgland, H. Penedones, S. Petersen, K. Simonyan, S. Crossan, P. Kohli, D. T. Jones, D. Silver, K. Kavukcuoglu, and D. Hassabis, *Improved Protein Structure Prediction Using Potentials from Deep Learning*, Nature **577**, 7792 (2020).
[6] B. Meredig, A. Agrawal, S. Kirklin, J. E. Saal, J. W. Doak, A. Thompson, K. Zhang, A. Choudhary, and C. Wolverton, *Combinatorial Screening for New Materials in Unconstrained Composition Space with Machine Learning*, Physical Review B **89**, 094104 (2014).
[7] J. Carrete, W. Li, N. Mingo, S. Wang, and S. Curtarolo, *Finding Unprecedentedly Low-Thermal-Conductivity Half-Heusler Semiconductors via High-Throughput Materials Modeling*, Physical Review X **4**, 011019 (2014).
[8] L. Bassman, P. Rajak, R. K. Kalia, A. Nakano, F. Sha, J. Sun, D. J. Singh, M. Aykol, P. Huck, and K. Persson, *Active Learning for Accelerated Design of Layered Materials*, Npj Computational Materials **4**, 1 (2018).
[9] K. Kaufmann, D. Maryanovsky, W. M. Mellor, C. Zhu, A. S. Rosengarten, T. J. Harrington, C. Oses, C. Toher, S. Curtarolo, and K. S. Vecchio, *Discovery of High-Entropy Ceramics via Machine Learning*, Npj Computational Materials **6**, 1 (2020).

[10] J. C. Snyder, M. Rupp, K. Hansen, K.-R. Müller, and K. Burke, *Finding Density Functionals with Machine Learning*, Phys. Rev. Lett. **108**, 253002 (2012).

[11] Z. Li, J. R. Kermode, and A. De Vita, *Molecular Dynamics with On-the-Fly Machine Learning of Quantum-Mechanical Forces*, Phys. Rev. Lett. **114**, 096405 (2015).

[12] J. Behler and M. Parrinello, *Generalized Neural-Network Representation of High-Dimensional Potential-Energy Surfaces*, Phys. Rev. Lett. **98**, 146401 (2007).

[13] T. L. Jacobsen, M. S. Jørgensen, and B. Hammer, *On-the-Fly Machine Learning of Atomic Potential in Density Functional Theory Structure Optimization*, Phys. Rev. Lett. **120**, 026102 (2018).

[14] M. Raissi, P. Perdikaris, and G. E. Karniadakis, *Physics Informed Deep Learning (Part I): Data-Driven Solutions of Nonlinear Partial Differential Equations*, ArXiv:1711.10561 [Cs, Math, Stat] (2017).

[15] M. Raissi, P. Perdikaris, and G. E. Karniadakis, *Physics Informed Deep Learning (Part II): Data-Driven Discovery of Nonlinear Partial Differential Equations*, ArXiv:1711.10566 [Cs, Math, Stat] (2017).

[16] J. Ling, A. Kurzawski, and J. Templeton, *Reynolds Averaged Turbulence Modelling Using Deep Neural Networks with Embedded Invariance*, Journal of Fluid Mechanics **807**, 155 (2016).

[17] M. Schmidt and H. Lipson, *Distilling Free-Form Natural Laws from Experimental Data*, Science **324**, 81 (2009).

[18] S. H. Rudy, S. L. Brunton, J. L. Proctor, and J. N. Kutz, *Data-Driven Discovery of Partial Differential Equations*, Science Advances **3**, e1602614 (2017).

[19] K. Champion, B. Lusch, J. N. Kutz, and S. L. Brunton, *Data-Driven Discovery of Coordinates and Governing Equations*, Proceedings of the National Academy of Sciences **116**, 22445 (2019).

[20] S. A. Harp, T. Samad, and A. Guha, *Designing Application-Specific Neural Networks Using the Genetic Algorithm*, in *Advances in Neural Information Processing Systems* (1990), pp. 447–454.

[21] G. F. Miller, P. M. Todd, and S. U. Hegde, *Designing Neural Networks Using Genetic Algorithms.*, in *ICGA*, Vol. 89 (1989), pp. 379–384.

[22] S. W. Stepniewski and A. J. Keane, *Pruning Backpropagation Neural Networks Using Modern Stochastic Optimisation Techniques*, Neural Computing & Applications **5**, 76 (1997).

[23] X. Yao and Y. Liu, *A New Evolutionary System for Evolving Artificial Neural Networks*, IEEE Transactions on Neural Networks **8**, 694 (1997).

[24] D. J. Montana and L. Davis, *Training Feedforward Neural Networks Using Genetic Algorithms.*, in *IJCAI*, Vol. 89 (1989), pp. 762–767.

[25] F. P. Such, V. Madhavan, E. Conti, J. Lehman, K. O. Stanley, and J. Clune, *Deep Neuroevolution: Genetic Algorithms Are a Competitive Alternative for Training Deep Neural Networks for Reinforcement Learning*, ArXiv Preprint ArXiv:1712.06567 (2017).

[26] A. Costa, R. Dangovski, S. Kim, P. Goyal, M. Soljačić, and J. Jacobson, *Interpretable Neuroevolutionary Models for Learning Non-Differentiable Functions and Programs*, ArXiv Preprint ArXiv:2007.10784 (2020).

[27] S. Greydanus, M. Dzamba, and J. Yosinski, *Hamiltonian Neural Networks*, ArXiv:1906.01563 [Cs] (2019).

[28] Z. Chen, J. Zhang, M. Arjovsky, and L. Bottou, *Symplectic Recurrent Neural Networks*, ArXiv:1909.13334 [Cs, Stat] (2019).

[29] M. J. Eslamibidgoli, M. Mokhtari, and M. H. Eikerling, *Recurrent Neural Network-Based Model for Accelerated Trajectory Analysis in AIMD Simulations*, ArXiv:1909.10124 [Physics] (2019).

[30] S.-M. Udrescu and M. Tegmark, *AI Feynman: A Physics-Inspired Method for Symbolic Regression*, Science Advances **6**, eaay2631 (2020).

[31] R. Iten, T. Metger, H. Wilming, L. Del Rio, and R. Renner, *Discovering Physical Concepts with Neural Networks*, Physical Review Letters **124**, 010508 (2020).

[32] G. Martius and C. H. Lampert, *Extrapolation and Learning Equations*, ArXiv Preprint ArXiv:1610.02995 (2016).

[33] S. Kim, P. Lu, S. Mukherjee, M. Gilbert, L. Jing, V. Ceperic, and M. Soljacic, *Integration of Neural Network-Based Symbolic Regression in Deep Learning for Scientific Discovery*, ArXiv Preprint ArXiv:1912.04825 (2019).

[34] H. Xu, H. Chang, and D. Zhang, *DLGA-PDE: Discovery of PDEs with Incomplete Candidate Library via Combination of Deep Learning and Genetic Algorithm*, Journal of Computational Physics 109584 (2020).

[35] F. A. Lindemann, *Uber Die Berechnung Molekularer Eigenfrequenzen*, Z. Physik **11**, 609 (1910).

[36] S. Desai and A. Strachan, *Discovering Discretized Classical Equations of Motion Using Parsimonious Neural Networks*, (2020).

[37] Z. Long, Y. Lu, and B. Dong, *PDE-Net 2.0: Learning PDEs from Data with a Numeric-Symbolic Hybrid Deep Network*, Journal of Computational Physics **399**, 108925 (2019).

[38] L. Verlet, *Computer "Experiments" on Classical Fluids. I. Thermodynamical Properties of Lennard-Jones Molecules*, Phys. Rev. **159**, 98 (1967).

[39] M. Tuckerman, B. J. Berne, and G. J. Martyna, *Reversible Multiple Time Scale Molecular Dynamics*, The Journal of Chemical Physics **97**, 1990 (1992).

[40] H. Yoshida, *Construction of Higher Order Symplectic Integrators*, Physics Letters A **150**, 262 (1990).

[41] G. Rowlands, *A Numerical Algorithm for Hamiltonian Systems*, Journal of Computational Physics **97**, 235 (1991).

[42] J. A. Izaguirre, S. Reich, and R. D. Skeel, *Longer Time Steps for Molecular Dynamics*, The Journal of Chemical Physics **110**, 9853 (1999).

[43] R. De Vogelaere, *Methods of Integration Which Preserve the Contact Transformation Property of the Hamilton Equations*, Technical Report (University of Notre Dame. Dept. of Mathematics) (1956).

[44] H. F. Trotter, *On the Product of Semi-Groups of Operators*, Proceedings of the American Mathematical Society **10**, 545 (1959).

[45] H. Eshet, R. Z. Khaliullin, T. D. Kühne, J. Behler, and M. Parrinello, *Ab Initio Quality Neural-Network Potential for Sodium*, Physical Review B **81**, 184107 (2010).

[46] J. Behler, R. Martoňák, D. Donadio, and M. Parrinello, *Metadynamics Simulations of the High-Pressure Phases of Silicon Employing a High-Dimensional Neural Network Potential*, Physical Review Letters **100**, 185501 (2008).

[47] S. Chmiela, H. E. Sauceda, K.-R. Müller, and A. Tkatchenko, *Towards Exact Molecular Dynamics Simulations with Machine-Learned Force Fields*, Nature Communications **9**, 1 (2018).

[48] F.-A. Fortin, F.-M. De Rainville, M.-A. Gardner, M. Parizeau, and C. Gagné, *DEAP: Evolutionary Algorithms Made Easy*, Journal of Machine Learning Research **13**, 2171 (2012).

[49] F. Chollet, *Keras* (2015).

[50] E. Hairer, C. Lubich, and G. Wanner, *Geometric Numerical Integration Illustrated by the Stormer-Verlet Method*, Acta Numerica **12**, 399 (2003).

[51] M. Born, *Thermodynamics of Crystals and Melting*, The Journal of Chemical Physics **7**, 591 (1939).

[52] T. Bäck, D. B. Fogel, and Z. Michalewicz, *Evolutionary Computation 1: Basic Algorithms and Operators* (CRC press, 2018).